\documentclass[preprint,5p,times,twocolumn]{elsarticle}

\usepackage{amssymb}
\usepackage{amsmath}
\usepackage{booktabs} 
\usepackage{algorithm}
\usepackage{algorithmic}
\usepackage{float}
\usepackage[colorlinks=true, linkcolor=blue, citecolor=blue, urlcolor=blue]{hyperref}
\usepackage{makecell} 
\usepackage{multirow}   
 
\journal{Information Sciences}

\begin{document}

\begin{frontmatter}



\title{\textsc{MedCoRAG}: Interpretable Hepatology Diagnosis via Hybrid Evidence Retrieval and Multispecialty Consensus} 

\author[label1]{Zheng Li}
\ead{lizheng050427@163.com}

\author[label1]{Jiayi Xu}
\ead{xujiayi041109@163.com}

\author[label1]{Zhikai Hu}
\ead{huzhikai1115@163.com}

\author[label2]{Hechang Chen}
\ead{chenhc@jlu.edu.cn}

\author[label3]{Lele Cong} 
\ead{congll18@mails.jlu.edu.cn}

\author[label4]{Yunyun Wang\corref{cor1}}
\ead{wangyunyun@jlu.edu.cn}

\author[label1,label5]{Shuchao Pang\corref{cor1}} 
    \ead{pangshuchao@njust.edu.cn}

\cortext[cor1]{Corresponding author}

\affiliation[label1]{organization={School of Cyber Science and Engineering, Nanjing University of Science and Technology},
            city={Nanjing},
            postcode={210094},
            country={China}}

\affiliation[label2]{organization={School of Artificial Intelligence, Jilin University},
            city={Changchun},
            postcode={130015},
            country={China}}

\affiliation[label3]{organization={Department of Neurology, China-Japan Union Hospital of Jilin University},
            city={Changchun},
            postcode={130033},
            country={China}}

\affiliation[label4]{organization={Department of Anesthesiology, China-Japan Union Hospital of Jilin University},
            city={Changchun},
            postcode={130033},
            country={China}}

\affiliation[label5]{organization={School of Computing, Macquarie University},
            city={Sydney},
            postcode={NSW 2109},
            country={Australia}}

\begin{abstract}
Diagnosing hepatic diseases accurately and interpretably is critical, yet it remains challenging in real-world clinical settings. Existing AI approaches for clinical diagnosis often lack transparency, structured reasoning, and deployability. Recent efforts have leveraged large language models (LLMs), retrieval-augmented generation (RAG), and multi-agent collaboration. However, these approaches typically retrieve evidence from a single source and fail to support iterative, role-specialized deliberation grounded in structured clinical data. To address this, we propose \textsc{MedCoRAG} (i.e., Medical Collaborative RAG), an end-to-end framework that generates diagnostic hypotheses from standardized abnormal findings and constructs a patient-specific evidence package by jointly retrieving and pruning UMLS knowledge graph paths and clinical guidelines. It then performs Multi-Agent Collaborative Reasoning: a Router Agent dynamically dispatches Specialist Agents based on case complexity; these agents iteratively reason over the evidence and trigger targeted re-retrievals when needed, while a Generalist Agent synthesizes all deliberations into a traceable consensus diagnosis that emulates multidisciplinary consultation. Experimental results on hepatic disease cases from MIMIC-IV show that \textsc{MedCoRAG} outperforms existing methods and closed-source models in both diagnostic performance and reasoning interpretability.
\end{abstract}



\begin{keyword}
Clinical Decision Support \sep Large Language Models\sep Retrieval-Augmented Generation \sep Multiple Agents


\end{keyword}

\end{frontmatter}



\section{Introduction}
\label{sec1}

In the era of AI-driven precision medicine, accurate and interpretable diagnosis of hepatic diseases from real-world Electronic Health Records (EHRs) is vital but challenging. Early detection, which enables timely intervention, is crucial to prevent irreversible damage and significantly improve outcomes. However, these conditions often present with vague, overlapping symptoms \cite{gan2025liver}. This clinical urgency demands not only reliable but also transparent methods that translate complex EHR data into actionable and explainable diagnostic insights.

Large language models have demonstrated impressive capabilities on general medical benchmarks \cite{singhal2025toward,tang2025medagentsbench} and offer a promising foundation for clinical AI. However, when deployed on real-world EHR data such as MIMIC-IV \cite{johnson2023mimic}, they face significant hurdles in the hepatology context: their knowledge is static and potentially outdated, sometimes yielding confident yet incorrect diagnoses \cite{griot2025large,hager2024evaluation,zhu2024clinicrealm}. More critically, their reasoning processes lack traceable, step-by-step justification—making it difficult to align model outputs with the interpretability standards required for high-stakes liver disease diagnosis.

Retrieval-augmented generation (RAG) \cite{gao2023retrieval} has emerged as a strategy to ground large language model outputs in external evidence. While traditional RAG relies on unstructured text, it struggles with multi-hop clinical inference due to the absence of explicit medical relationships. Recent approaches integrate medical knowledge graphs (KGs) \cite{hu2025graggraphretrievalaugmentedgeneration,wu2024medical} to enable structured reasoning, significantly outperforming classic RAG by leveraging semantic paths between concepts. However, raw KG paths often contain irrelevant or implausible links, and these methods still fail to incorporate context-sensitive guidance from clinical practice guidelines—such as diagnostic criteria or evidence hierarchies. Meanwhile, emerging multi-agent frameworks aim to emulate multidisciplinary consultation by deploying specialized agents that collaboratively debate differentials through iterative dialogue \cite{tang2025medagentsbench,tang2024medagents,kim2024mdagents}. Yet most systems \cite{wang2025lins,wang2025colacare} operate over loosely retrieved or internal knowledge, lacking deep integration of both KGs and authoritative guidelines, and thus fall short of evidence-based standards. Moreover, agent activation is typically static, with the same set of specialists engaged regardless of case complexity \cite{schmidgall2024agentclinic,kim2024mdagents}, resulting in either redundant deliberation or insufficient expertise. Notably, even advanced agent-based systems often prioritize end-task accuracy over interpretable, clinician-aligned diagnostic reasoning, limiting trust in complex hepatobiliary decisions \cite{hellingman2020value,smith2025acr,hong2024argmed}.

\begin{figure}
    \centering
    \includegraphics[width=1.0\linewidth]{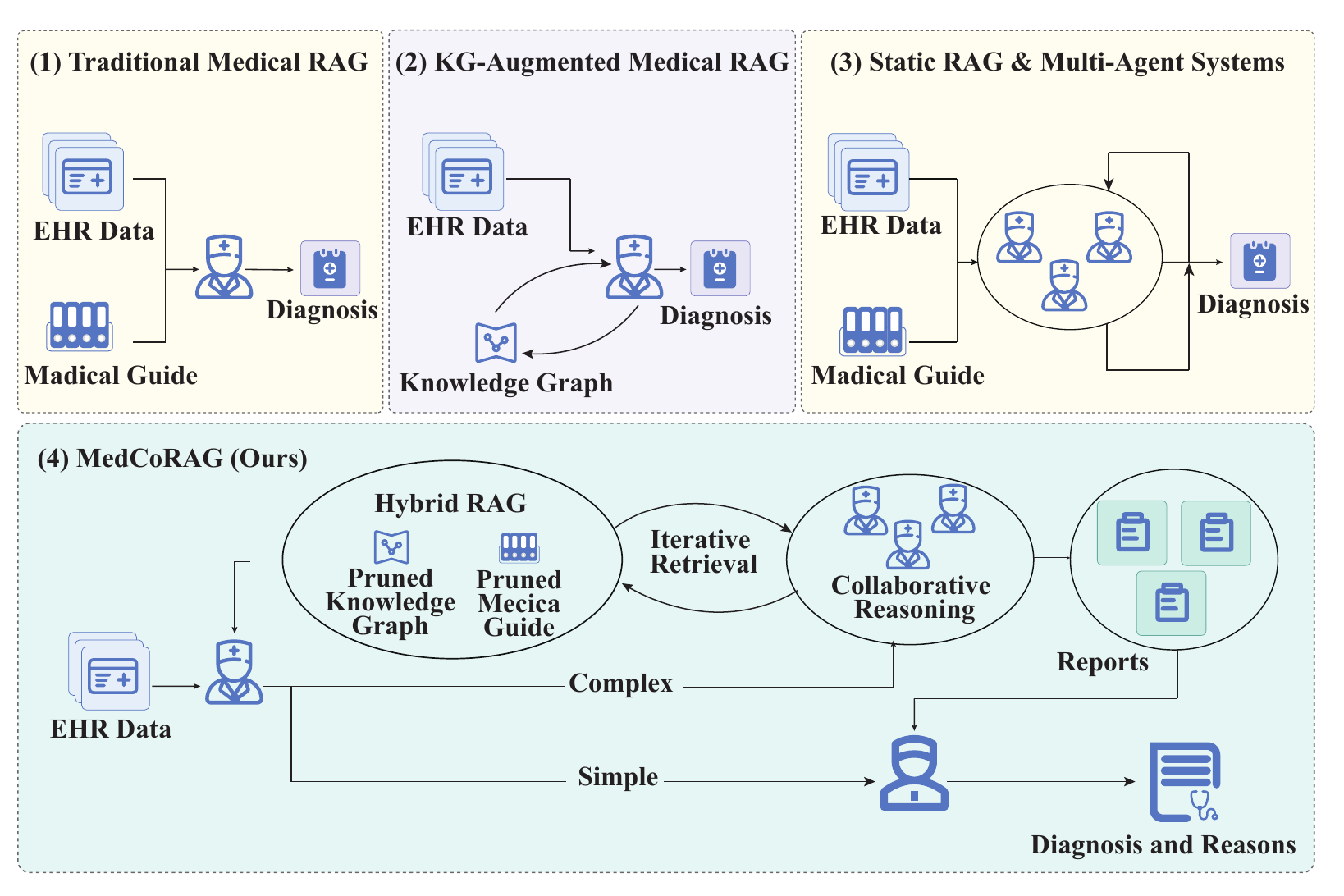}
    \caption{Comparative Overview of Medical Diagnostic Reasoning Frameworks}
    \label{fig:Comparative Overview of Medical Diagnostic Reasoning Frameworks}
\end{figure}

To address these gaps, we propose \textsc{MedCoRAG} (i.e., Medical Collaborative RAG), a hybrid RAG and multi-agent framework that grounds multidisciplinary clinical reasoning \cite{pillay2016impact} in unified evidence synthesis. As shown in Figure~\ref{fig:Comparative Overview of Medical Diagnostic Reasoning Frameworks}, unlike prior approaches that rely on static agent teams or limited sources of external evidence, \textsc{MedCoRAG} integrates guideline-constrained KG pruning with dynamic, complexity-aware specialist dispatch. We first transform structured EHR data into coherent clinical narratives and generate an initial set of diagnostic hypotheses. For each hypothesis, the system retrieves clinical guideline excerpts \cite{european2025easl,taddei2025critical,arab2025metabolic} and UMLS \cite{Bodenreider2004TheUM} knowledge graph paths, then prunes the paths using an LLM that evaluates their clinical coherence against the full narrative and guidelines, yielding a patient-tailored evidence package for all cases. A Router Agent assesses case complexity based on the clinical narrative and abnormal findings. In simple cases, a Generalist Agent directly synthesizes a diagnosis based on the initial evidence package. In complex cases, the system dynamically dispatches relevant specialist agents such as Hepatology or Oncology based on clinical context; these specialists perform iterative reasoning over the evidence package and trigger targeted retrievals whenever current evidence is insufficient. The Generalist Agent ultimately produces a single, traceable consensus diagnosis through holistic adjudication of all specialist inputs, deliberation history, and unresolved uncertainties. 

We evaluate \textsc{MedCoRAG} on real-world hepatic cases from MIMIC-IV and show that it generates precise and evidence-grounded diagnoses. The framework’s dynamic routing and evidence-pruning mechanisms suppress spurious associations and redundant deliberation, yielding focused and interpretable reasoning.

To summarize, our contributions are as follows:
\begin{itemize}

  \item We propose \textsc{MedCoRAG}, a multi-specialty RAG–agent framework that dynamically emulates hepatology MDT consultations by coordinating on-demand specialist agents in an iterative, evidence-constrained diagnostic loop over shared, guideline-pruned multi-hop knowledge graph paths.

  \item We introduce MDT-aligned hybrid reasoning, a method to unify pruned KG paths and clinical guideline excerpts into a single evidential space that is jointly interpreted through role-specific specialist lenses, yielding interpretable and hallucination-resistant consensus diagnoses grounded in real-world clinical practice.
    
\item We conduct comprehensive experiments on hepatic disease cases from the MIMIC-IV dataset. Experimental results demonstrate the effectiveness of \textsc{MedCoRAG}, which achieves high performance across various diagnostic metrics.

\end{itemize}

\section{Related Work}
\label{sec:related_work}
\paragraph{Medical Retrieval and Knowledge-Augmented Reasoning}
Standard RAG mitigates LLM hallucinations by grounding responses in medical literature or EHRs \cite{amugongo2025retrieval}, but its similarity-based retrieval often returns irrelevant passages and fails to support iterative clinical reasoning \cite{xiong2024benchmarking}. Recent efforts address this in complementary ways: MedGraphRAG \cite{wu-etal-2025-medical} constructs a multi-tier knowledge graph from academic papers, medical dictionaries and clinical guidelines, using hierarchical clustering to generate structured tag summaries for coherent retrieval; MedRAG \cite{zhao2025medrag} improves diagnostic alignment by grouping diseases via symptom similarity rather than diagnostic codes; KG-Rank \cite{yang2024kg} boosts answer quality by re-ranking retrieved passages using KG-derived entity-path relevance—achieving over 18\% gain in ROUGE-L without modifying the LLM; and rationale-guided RAG \cite{sohn2025rationale} first generates a lightweight diagnostic rationale to steer single-step retrieval, improving precision without fine-tuning. Despite these advances, each approach operates in isolation—none jointly integrates structured KG paths, full clinical guidelines, and adaptive retrieval. Our method bridges this gap by jointly retrieving KG-derived reasoning paths and guideline excerpts, then applying domain-aware pruning to produce a focused, traceable, and clinically coherent evidence package.

\paragraph{Multi-Agent Systems for Clinical Collaboration}
Early multi-agent frameworks established role-based clinical collaboration through zero-shot role playing \cite{tang2024medagents}, argumentation-driven explainability \cite{hong2024argmed}, or simulated clinical environments \cite{schmidgall2024agentclinic,li2024agent}. Recent work shifts toward evidence-grounded and optimized workflows: ColaCare \cite{wang2025colacare} and LINS \cite{wang2025lins} coordinate agents over structured EHRs or citation-backed chains; TxAgent \cite{gao2025txagent} focuses on dynamic tool composition for therapy planning; and MedAide \cite{yang2025medaide} fuses intent-aware extractors for multifaceted reasoning. Notably, MedAgent-Pro \cite{wang2025medagent} introduces a reasoning agentic workflow that constructs traceable diagnostic paths from multimodal inputs and clinical guidelines, while MMedAgent-RL \cite{xia2025mmedagent} leverages reinforcement learning to optimize agent collaboration policies for improved diagnostic accuracy. Despite these advances, most systems either fix collaboration structures or decouple deliberation from a unified, pruned evidence base—limiting adaptability and clinical fidelity. Our approach addresses this by dynamically routing specialists based on abnormal findings and coordinating their RL-informed deliberation over a shared, guideline-anchored knowledge graph.

\section{Methodology}
\subsection{Overall Architecture}
\label{subsec:overall_architecture}

\begin{figure*}
    \centering
    \includegraphics[width=1\linewidth]{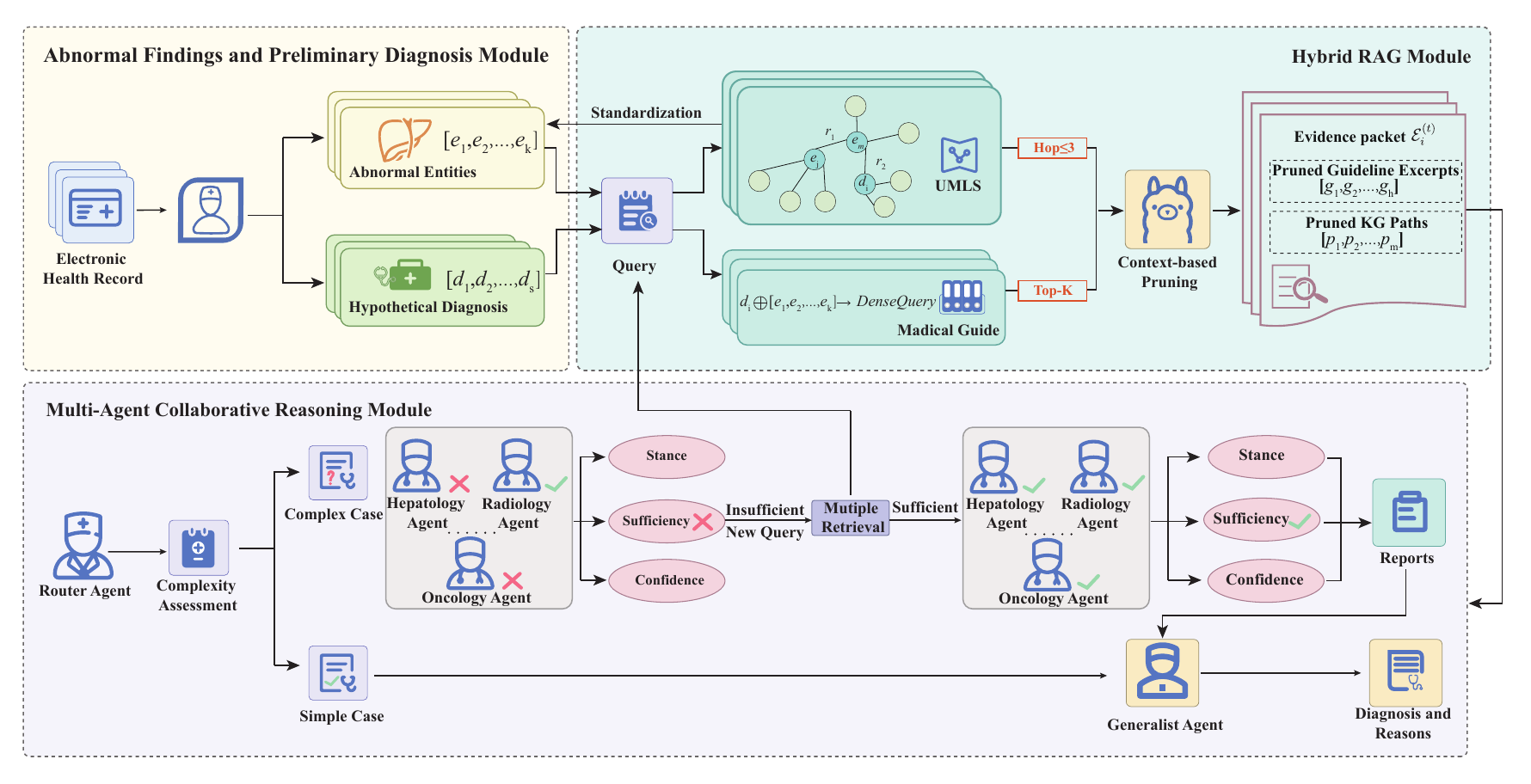}
\caption{Overall architecture of the \textsc{MedCoRAG} framework, comprising three core components. (1) Abnormal Findings and Preliminary Diagnosis: Abnormal clinical findings are extracted from the patient narrative and standardized via UMLS to generate a focused set of initial diagnostic hypotheses. (2) Hybrid RAG: For each hypothesis, the system retrieves clinical guideline excerpts and UMLS knowledge graph paths, then prunes them using the full clinical context to form a coherent, patient-specific evidence package. (3) Multi-Agent Collaborative Reasoning: A Router Agent assesses case complexity to either activate relevant specialist agents or delegate simple cases to the Generalist Agent; all agents iteratively reason over the shared evidence, trigger re-retrieval when needed, and converge on an interpretable consensus diagnosis through the Generalist Agent.}
    \label{fig:Overall architecture}
\end{figure*}

\textsc{MedCoRAG} implements an end-to-end diagnostic workflow grounded in structured evidence synthesis, as shown in Figure~\ref{fig:Overall architecture}. From the initial set of diagnostic hypotheses, the system performs a first-round retrieval of clinical guideline excerpts and multi-hop knowledge graph paths, which are jointly pruned using the full clinical narrative to form patient-specific evidence packages. A Router Agent then assesses case complexity based on the narrative and abnormal findings: for simple cases, a Generalist Agent directly renders a diagnosis from the initial evidence; for complex cases, it dynamically dispatches specialty-specific agents. These specialists iteratively evaluate hypotheses against the shared evidence, triggering agent-guided re-retrieval when needed. The final diagnosis is produced by the Generalist through holistic adjudication of all deliberations—yielding a single, traceable, and clinically actionable conclusion.

\subsection{Core Components}
\subsubsection{Abnormal Findings and Preliminary Diagnosis}
\label{subsubsec:hypothesis_generation}

\paragraph{Abnormal Entity Recognition and Standardization}
Given a patient’s case description $\mathcal{C}$, the system first invokes an LLM to extract candidate abnormal entities:
\begin{equation}
    \mathcal{E}_{\text{raw}} = \text{LLM}_{\text{NER}}(\mathcal{C}),
\end{equation}
where $\text{LLM}_{\text{NER}}$ denotes a large language model prompted to identify diagnostic-relevant abnormalities from clinical text.

To align these entities with standardized medical terminology, the system queries the knowledge graph UMLS for each raw entity $e \in \mathcal{E}_{\text{raw}}$, obtaining a list of candidate standardized entities:
\begin{equation}
    \mathcal{S}(e) = \mathrm{KMatch}(e),
\end{equation}
where $\mathrm{KMatch}(\cdot)$ denotes knowledge graph entity matching.

The raw entity $e$ and its candidate matches $\mathcal{S}(e)$ are then presented to the LLM, which selects the best-matching standardized entity—or indicates no match:
\begin{equation}
    e_{\text{std}} = \text{LLM}_{\text{align}}\big( e, \mathcal{S}(e) \big) \in \mathcal{S}(e) \cup \{ \varnothing \}.
\end{equation}

The final set of standardized abnormal entities is defined as:
\begin{equation}
    \mathcal{E}_{\text{abn}} = \{ e_{\text{std}} \mid e \in \mathcal{E}_{\text{raw}},\ e_{\text{std}} \neq \varnothing \}.
\end{equation}

This ensures semantic consistency with the underlying knowledge graph, enabling precise downstream evidence retrieval.

\paragraph{Direct Generation of Candidate Diagnoses}
The system generates the initial diagnostic hypotheses using both the complete case description $\mathcal{C}$ and the standardized abnormal findings $\mathcal{E}_{\text{abn}}$. These inputs are formatted into a structured clinical prompt, and the LLM directly produces a concise list of plausible differential diagnoses.

Formally, the initial hypothesis list is generated as
\begin{equation}
    \mathcal{H}_{\text{initial}} = \text{LLM}_{\text{hypo}}(\mathcal{C}, \mathcal{E}_{\text{abn}}).
\end{equation}
The output takes the form of a bounded sequence $[d_1, d_2, \dots, d_K]$ with $K \leq K_{\max}$ ($K_{\max}=4$) to maintain diagnostic focus, where each $d_i$ is a standardized disease name.

\subsubsection{Hybrid RAG}
\label{subsubsec:hybrid_retrieval}

This module treats each candidate diagnosis $d_i$ as an anchor and retrieves concurrently two complementary evidence types: (1) authoritative statements from a clinical guideline corpus, and (2) interpretable reasoning paths from the UMLS knowledge graph. 

\paragraph{Clinical Guideline Retrieval and Relevance Filtering}
For each candidate diagnosis $d_i$, a composite query is constructed by combining $d_i$ with all abnormal findings in $\mathcal{E}_{\text{abn}}$, forming a semantically enriched representation that captures the clinical context of the hypothesis. This query drives a two-stage retrieval process over a pre-indexed clinical guideline database.

In the first stage, a bi-encoder computes dense embeddings for queries and guideline segments, and uses cosine similarity to retrieve the Top-$K$ most relevant segments ($K=8$). In the second stage, a cross-encoder re-ranks these candidates based on contextual alignment and selects the Top-$N$ segments ($N=4$). The resulting set is denoted as:
\begin{equation}
    \mathcal{G}_i = \mathrm{GRet}(d_i, \mathcal{E}_{\text{abn}}; N=4),
\end{equation}
where $\mathrm{GRet}(\cdot)$ denotes the two-stage guideline retrieval pipeline.

\paragraph{Knowledge Graph Path Retrieval and Guideline-Informed Pruning}
For each pair $(d_i, e_j)$ with $e_j \in \mathcal{E}_{\text{abn}}$, the system queries the UMLS knowledge graph for semantic paths from $e_j$ to $d_i$ with at most 3 hops:
\begin{equation}
    \mathcal{P}_{ij} = \mathrm{KRet}(e_j, d_i; h_{\max}=3),
\end{equation}
where $\mathrm{KRet}(\cdot)$ denotes knowledge graph path retrieval.

Raw paths may contain irrelevant semantic links. Each path $p$ is first verbalized by LLM into a natural-language statement linking $e_j$ to $d_i$ via intermediate concepts. To assess relevance, verbalized paths for $d_i$ (denoted $\mathcal{P}_i = \bigcup_{e_j \in \mathcal{E}_{\text{abn}}} \mathcal{P}_{ij}$) are batched (8 per batch). An LLM evaluates each batch using only the full case description $\mathcal{C}$ and the top guideline excerpts $\mathcal{G}_i^{\text{top}} \subseteq \mathcal{G}_i$, producing a binary judgment. Let $\phi_{\text{rel}}(p; \mathcal{C}, \mathcal{G}_i^{\text{top}}) \in \{0,1\}$ indicate whether path $p$ provides a clinically coherent and guideline-supported explanation for the patient’s presentation. Then:
\begin{equation}
    \mathcal{P}_i^{\text{valid}} = \left\{ p \in \mathcal{P}_i \,\middle|\, \phi_{\text{rel}}(p; \mathcal{C}, \mathcal{G}_i^{\text{top}}) = 1 \right\}.
\end{equation}

Finally, for each $d_i$, the system aggregates $\mathcal{P}_i^{\text{valid}}$ and $\mathcal{G}_i$ into a structured evidence package $\mathcal{E}_i^{(0)}$, which supports subsequent multi-agent reasoning.

\subsubsection{Multi-Agent Collaborative Reasoning}
\label{subsubsec:multi_agent_reasoning}

\paragraph{Complexity Assessment}
The Router Agent processes the complete patient case description $\mathcal{C}$—including free-text clinical history, physical examination notes, and narrative laboratory or imaging reports—together with standardized abnormal findings $\mathcal{E}_{\text{abn}}$. This holistic representation enables the detection of subtle indicators of diagnostic complexity, such as contradictory findings, multi-organ involvement, atypical presentations, and diagnostic uncertainty. Based on this integrated context, the Router applies a narrative-aware complexity discrimination function $\phi_{\text{comp}}$ to yield a binary decision:
\begin{equation}
    c = \phi_{\text{comp}}\left( \mathcal{C},\ \mathcal{E}_{\text{abn}},\ \mathcal{H}_{\text{initial}} \right),
\end{equation}
where $c \in \{0, 1\}$. When $c = 0$, the case is routed to a Generalist Agent, which synthesizes a final diagnosis directly from the pre-retrieved evidence package $\mathcal{E}_i^{(0)}$, bypassing multi-agent deliberation. When $c = 1$, the system initiates dynamic specialist dispatch for collaborative reasoning.

For complex cases ($c = 1$), the system performs \textit{dynamic specialist dispatch}.Instead of relying on a fixed specialty set, the system dynamically selects relevant specialists based on the clinical context. Specifically, a scheduling function $\psi$ identifies a contextually appropriate subset of specialists by jointly analyzing the semantic types of the abnormal findings and the full clinical narrative:
\begin{equation}
    \mathcal{A}_i = \psi\left( \mathcal{C},\ \mathcal{E}_{\text{abn}},\ d_i \right),
\end{equation}
where $\mathcal{A}_i$ denotes the set of specialists activated for evaluating diagnosis $d_i$.
For instance, a case describing ``jaundice, elevated creatinine, and maculopapular rash after drug initiation'' would activate Hepatology, Nephrology, and Dermatology agents. To enable efficient deployment, we employ knowledge distillation to transfer the clinical reasoning capabilities of a large model (Qwen3-Max \cite{yang2025qwen3}) into a student model (Llama-3.1-8B-Instruct \cite{dubey2024llama}), which powers the specialist agents during inference.

\paragraph{Evidence-Driven Specialist Reasoning}
Each dispatched specialist agent $a \in \mathcal{A}_i$ receives an identical reasoning context:
\begin{equation}
    \text{Context}_i^{(t)} = \left( \mathcal{C},\ \mathcal{E}_{\text{abn}},\ d_i,\ \mathcal{E}_i^{(t)} \right),
\end{equation}
where $\mathcal{E}_i^{(t)}$ denotes the evidence package at iteration $t$ (initialized at $t=0$), comprising knowledge graph paths and guideline excerpts retrieved via the hybrid RAG pipeline described in Section~\ref{subsubsec:hybrid_retrieval}. Guided by role-specific prompts, each agent independently assesses candidate diagnosis $d_i$, producing a stance $o_a^{(i)} \in \{\text{S}, \text{N}, \text{O}\}$ (support, neutral, oppose), a confidence score $c_a^{(i)} \in [0,1]$, and an evidence sufficiency judgment $s_a^{(i)} \in \{\text{Suf}, \text{Ins}\}$, accompanied by a justification explicitly anchored to items in $\mathcal{E}_i^{(t)}$.

If any agent reports $s_a^{(i)} = \text{Ins}$, the Coordinator Agent aggregates these signals to decide whether to initiate an additional retrieval round. Specifically, let 
\begin{equation}
    \rho^{(t)} = \frac{1}{|\mathcal{A}_i|} \sum_{a \in \mathcal{A}_i} \mathbb{I}\big(s_a^{(i)} = \text{Ins}\big)
\end{equation}
denote the proportion of agents deeming the current evidence insufficient. When $\rho^{(t)} > \tau_{\text{suff}}$ ($\tau_{\text{suff}} = 0.5$) and $t < T_{\max}$ ($T_{\max} = 3$), the system launches a targeted secondary retrieval: agents collaboratively formulate queries that address specific diagnostic uncertainties, such as requests for imaging characteristics or guideline criteria tied to biomarker thresholds; these queries are combined with $\mathcal{C}$, $\mathcal{E}_{\text{abn}}$, $d_i$, and identified knowledge gaps to construct refined retrieval inputs; the hybrid retrieval pipeline (Section~\ref{subsubsec:hybrid_retrieval}) is then re-executed to fetch supplementary knowledge graph paths and guideline excerpts; finally, the new evidence is merged into an updated package $\mathcal{E}_i^{(t+1)}$ for the next reasoning iteration. This closed-loop mechanism allows retrieval and deliberation to co-evolve until diagnostic confidence stabilizes or the iteration limit is reached.

\paragraph{Consensus Formation and Confidence Calibration}
At each iteration $t$, a Generalist Agent synthesizes all specialist inputs—including stances, confidence scores, evidence-based justifications, and sufficiency judgments—to generate an interim consensus report for each candidate diagnosis $d_i$. The calibrated confidence score $s_i^{(t)}$, which quantifies diagnostic plausibility, is computed as:
\begin{equation}
    s_i^{(t)} = \frac{1}{|\mathcal{A}_i|} \sum_{a \in \mathcal{A}_i} \mathbb{I}\big(o_a^{(i)} = \text{S}\big),
\end{equation}
where $\mathbb{I}(\cdot)$ denotes the indicator function. The process terminates early if $s_i^{(t)} > \tau_{\text{high}}$ ($\tau_{\text{high}} = 0.9$), signaling strong collective support for $d_i$.

\paragraph{Final Diagnosis Selection and Output}
Rather than selecting the diagnosis with the highest calibrated score alone, the Generalist Agent conducts a holistic final adjudication. It integrates the complete deliberation history—including evolving evidence packages, inter-agent agreements and disagreements, unresolved uncertainties, and patient-specific context ($\mathcal{C}$, $\mathcal{E}_{\text{abn}}$)—to produce a single, clinically coherent final diagnosis:
\begin{equation}
    d_{\text{final}} = \mathcal{F}_{\text{final}} \left( \left\{ (d_i, s_i^{(T)}, \text{Report}_i) \right\}_{d_i \in \mathcal{H}_{\text{initial}}} \right),
\end{equation}
where $\mathcal{F}_{\text{final}}$ denotes the Generalist’s adjudication function. This decision is accompanied by a comprehensive, traceable consensus report that synthesizes supporting and contradicting evidence across all iterations, highlights key clinical uncertainties, and recommends actionable next-step investigations.

The whole algorithm of \textsc{MedCoRAG} is shown in Algorithm ~\ref{alg:MedCoRAG}.

\begin{algorithm}[t]
\caption{\textsc{MedCoRAG} Diagnostic Workflow}
\label{alg:MedCoRAG}
\begin{algorithmic}[1]
\REQUIRE case description $\mathcal{C}$
\ENSURE Final diagnosis $d_{\text{final}}$ with justification

\STATE $\mathcal{E}_{\text{abn}} = \left\{ \text{LLM}_{\text{align}}\big(e, \mathrm{KMatch}(e)\big) \mid e \in \text{LLM}_{\text{NER}}(\mathcal{C}) \right\}$
\STATE $\mathcal{H} = \text{LLM}_{\text{hypo}}(\mathcal{C}, \mathcal{E}_{\text{abn}})$ \COMMENT{Top-$K$ focused hypotheses ($K \leq K_{\max}$)}

\FOR{each $d_i \in \mathcal{H}$}
    \STATE $\mathcal{G}_i = \mathrm{GRet}(d_i, \mathcal{E}_{\text{abn}}; N=4)$ 
    \STATE $\mathcal{P}_i^{\text{valid}} = \left\{ p \in \mathrm{KRet}(e_j, d_i; h_{\max}=3) \,\middle|\, \text{LLM}_{\text{prune}}(p; \mathcal{C}, \mathcal{G}_i^{\text{top}}) = 1 \right\}$ \COMMENT{for all $e_j \in \mathcal{E}_{\text{abn}}$}
    \STATE $\mathcal{E}_i^{(0)} \gets (\mathcal{G}_i, \mathcal{P}_i^{\text{valid}})$
\ENDFOR

\IF{$\phi_{\text{comp}}(\mathcal{C}, \mathcal{E}_{\text{abn}}, \mathcal{H}) = 0$}
    \STATE $d_{\text{final}} \gets \text{GeneralistAgent}(\mathcal{C}, \mathcal{E}_{\text{abn}}, \mathcal{E}_i^{(0)})$
    \RETURN $d_{\text{final}}$ with justification
\ENDIF

\FOR{$t = 1$ to $T_{\max}$}
    \STATE Specialists $a \in \mathcal{A}_i = \psi(\mathcal{C}, \mathcal{E}_{\text{abn}}, d_i)$ output $o_a^{(i)} \in \{\text{S}, \text{N}, \text{O}\}$ and $s_a^{(i)} \in \{\text{Suf}, \text{Ins}\}$
    \STATE Compute $\rho^{(t)} = \frac{1}{|\mathcal{A}_i|} \sum_{a} \mathbb{I}(s_a^{(i)} = \text{Ins})$
    \IF{$\rho^{(t)} \leq \tau_{\text{suff}}$}
        \STATE \textbf{break}
    \ELSE
        \STATE Update $\mathcal{E}_i^{(t+1)}$ via agent-proposed queries and hybrid retrieval
    \ENDIF
\ENDFOR

\STATE $d_{\text{final}} = \mathcal{F}_{\text{final}}\big( \{ (d_i, \text{Report}_i) \}_{i} \big)$
\RETURN $d_{\text{final}}$ with traceable consensus report
\end{algorithmic}
\end{algorithm}

\section{Experiments}
\label{sec:experiments}

We conduct a comprehensive evaluation of \textsc{MedCoRAG} on a real-world hepatic disease diagnosis task to assess the effectiveness of our abnormal-entity-driven reasoning framework, multi-source knowledge integration, and multi-agent collaboration mechanism.

\subsection{Dataset}
\label{subsec:dataset}
We curate a clinical dataset from the public MIMIC-IV database~\cite{johnson2023mimic}, focusing on patients diagnosed with one of 13 common hepatic diseases, whose standardized abbreviations are listed in Table~\ref{tab:disease_abbreviations}. To reflect the chronic and progressive nature of hepatic conditions, we retain all hospital admissions per patient, thus we reconstruct longitudinal medical histories. All data are fully de-identified.

Since MIMIC-IV primarily provides structured tabular records lacking the narrative context, we synthesize realistic, context-rich clinical narratives from each patient’s longitudinal timeline using LLM. These narratives are then formatted as medical question–answering pairs. The final dataset contains 3470 QA samples, split into training and test sets at a 7:3 ratio during the distillation phase, stratified by disease category to ensure balanced representation across all 13 classes.

\begin{table}[t]
\small
\centering
\caption{Disease abbreviations}
\label{tab:disease_abbreviations}
\begin{tabular}{@{}c c@{}}
\toprule
\textbf{Hepatic Disease} & \textbf{Abbreviation} \\
\midrule
Hepatitis B & HBV \\
Primary biliary cholangitis & PBC \\
Secondary liver cancer & SLC \\
Liver cyst & LCyst \\
Hepatoblastoma & HB \\
Liver cirrhosis & LC \\
Hepatocellular carcinoma & HCC \\
Hepatic hemangioma & HH \\
Liver failure & LF \\
Autoimmune hepatitis & AIH \\
Drug-induced liver injury & DILI \\
Non-alcoholic steatohepatitis & NASH \\
Rupture and bleeding of esophagogastric varices & EGVB \\
\bottomrule
\end{tabular}
\end{table}

\subsection{Evaluation Metrics}
We evaluate diagnostic performance using four standard metrics: Recall, Precision, F1-score, and F0.5-score. All metrics are reported as weighted averages across the 13 hepatic disease classes to account for class imbalance. 

\subsection{Baseline Models}
\label{subsec:baselines}

We compare \textsc{MedCoRAG} against a comprehensive set of baselines spanning model scale and reasoning architecture, all evaluated on the same test set using identical clinical narratives. This includes medical-domain models with up to 8B parameters—Qwen3-Medical-GRPO-4B~\cite{tang_2025_qwen3_medical_grpo}, OpenBioLLM-Llama3-8B~\cite{OpenBioLLMs}, Bio-Medical-Llama3-8B~\cite{ContactDoctor_Bio_Medical_Llama_3_8B}, and Llama3-Med42-8B~\cite{med42v2}; large proprietary models including DeepSeek-V3.1-Think~\cite{deepseekai2024deepseekv3technicalreport}, Gemini-2.5-Pro~\cite{comanici2025gemini}, GLM-4.6~\cite{zeng2025glm}, and GPT-4o~\cite{openai2024gpt4ocard}; medium-sized  models ranging from 14B to 32B parameters, including DeepSeek-R1-Distill-Qwen-32B~\cite{deepseekai2025deepseekr1incentivizingreasoningcapability}, GPT-OSS-20B~\cite{openai2025gptoss120bgptoss20bmodel}, Gemma3-27B~\cite{gemmateam2025gemma3technicalreport}, Qwen-QWQ-32B~\cite{qwq32b}, and Phi-4-14B~\cite{abdin2024phi}; lightweight models under 7B parameters, namely ChatGLM3-6B~\cite{glm2024chatglm} and Mistral-7B~\cite{jiang2023mistral7b}; recent RAG \& multi-agent diagnostic frameworks —ColaCare~\cite{wang2025colacare}, MedAgent-Pro~\cite{wang2025medagent}, and MedAide~\cite{yang2025medaide}— implemented using the same base model as \textsc{MedCoRAG} (Llama-3.1-8B-Instruct~\cite{dubey2024llama}). All methods are evaluated under the same protocol.

\subsection{Implementation Details}
\label{subsec:settings}

For retrieval-augmented reasoning, we integrate two complementary external knowledge sources: (1) a structured biomedical knowledge graph built upon the UMLS, and (2) an unstructured corpus of 38 authoritative clinical guidelines on hepatic diseases issued by major professional societies such as AASLD, EASL, APASL. For clinical guidelines, we perform dense retrieval using Qwen3-Embedding-8B \cite{qwen3embedding}, with embeddings indexed in Milvus \cite{wang2021milvus}, followed by re-ranking with Qwen3-Reranker-8B \cite{qwen3embedding}.

We use Llama-3.1-8B-Instruct as the backbone LLM. To facilitate deployment, we distill the reasoning capability of Qwen3-Max~\cite{yang2025qwen3} into Llama-3.1-8B-Instruct. Specifically, the teacher model simulates specialist agent behavior within the \textsc{MedCoRAG} framework to generate training data. The student model is fine-tuned via supervised learning with LoRA~\cite{hu2022lora} for three epochs, using a cosine-decayed learning rate initialized at $5 \times 10^{-5}$, an effective batch size of 8 via gradient accumulation, and a maximum sequence length of 11,000 tokens, all in \texttt{bf16} precision on a single A800 GPU.

\subsection{Main Results}
\label{subsec:results}

\begin{table*}[t]
\centering
\caption{Comprehensive diagnostic performance (\%) across model categories. The best result in each metric is \textbf{bolded}.}
\label{tab:main_results}
\begin{tabular}{@{}l l c c c c@{}}
\toprule
\textbf{Category} & \textbf{Model} & \textbf{Precision} & \textbf{Recall} & \textbf{F1-score} & \textbf{F0.5-score} \\
\midrule
\multirow{4}{*}{Medical ($\leq$8B)} 
 & Qwen3-Medical-GRPO-4B \cite{tang_2025_qwen3_medical_grpo} & 69.68 & 58.93 & 61.07 & 59.06 \\
 & OpenBioLLM-Llama3-8B \cite{OpenBioLLMs} & 62.15 & 51.48 & 54.21 & 52.24 \\
 & Bio-Medical-Llama3-8B \cite{ContactDoctor_Bio_Medical_Llama_3_8B} & 65.94 & 41.83 & 47.43 & 43.36 \\
 & Llama3-Med42-8B \cite{med42v2} & 64.57 & 61.41 & 60.28 & 60.51 \\
\midrule
\multirow{4}{*}{General large (>100B)} 
 & DeepSeek-V3.1-Think \cite{deepseekai2024deepseekv3technicalreport} & 79.59 & 76.98 & 77.61 & 77.05 \\
 & Gemini-2.5-Pro \cite{comanici2025gemini} & 80.31 & 76.70 & 77.28 & 76.71 \\
 & GLM-4.6 \cite{zeng2025glm} & 80.76 & 75.36 & 76.55 & 75.53 \\
 & GPT-4o \cite{openai2024gpt4ocard} & 74.33 & 70.58 & 69.98 & 69.90 \\
\midrule
\multirow{5}{*}{General medium (14--32B)} 
 & DeepSeek-R1-Distill-Qwen-32B \cite{deepseekai2025deepseekr1incentivizingreasoningcapability} & 78.49 & 74.59 & 74.90 & 74.35 \\
 & GPT-OSS-20B \cite{openai2025gptoss120bgptoss20bmodel} & 75.19 & 69.25 & 70.75 & 69.56 \\
 & Gemma3-27B \cite{gemmateam2025gemma3technicalreport} & 73.89 & 60.84 & 65.92 & 62.61 \\
 & Qwen-QWQ-32B \cite{qwq32b} & 66.60 & 53.96 & 55.78 & 54.02 \\
 & Phi-4-14B \cite{abdin2024phi} & 62.70 & 54.44 & 55.68 & 54.26 \\
\midrule
\multirow{2}{*}{General small ($\leq$7B)} 
 & ChatGLM3-6B \cite{glm2024chatglm} & 56.57 & 40.97 & 41.97 & 40.89 \\
 & Mistral-7B \cite{jiang2023mistral7b} & 60.86 & 40.02 & 41.52 & 39.12 \\
\midrule
\multirow{3}{*}{Agent Frameworks} 
 & ColaCare \cite{wang2025colacare} & 78.01 & 72.66 & 73.35 & 72.53 \\
 & MedAgent-Pro \cite{wang2025medagent} & 76.01 & 70.58 & 70.33 & 69.97 \\
 & MedAide \cite{yang2025medaide} & 77.98 & 73.23 & 74.10 & 73.26 \\
\midrule
\textbf{Ours} 
 & \textsc{MedCoRAG} & \textbf{81.32} & \textbf{79.18} & \textbf{79.12} & \textbf{78.99} \\
\bottomrule
\end{tabular}
\end{table*}

\label{subsec:analysis}
We conduct diagnostic classification across 13 hepatic disease categories on a standardized clinical test set. The overall performance of \textsc{MedCoRAG} is shown in Table~\ref{tab:main_results}. Among all evaluated methods, \textsc{MedCoRAG} achieves the best weighted Precision, Recall, F1-score, and F0.5-score. It performs better than specialized medical models with up to 8B parameters, large proprietary language models, medium- and small-scale general-purpose models, as well as recent multi-agent diagnostic frameworks. These results indicate that \textsc{MedCoRAG}’s approach—combining structured evidence synthesis with dynamic agent collaboration—can support accurate and reliable clinical diagnosis across diverse model scales and reasoning strategies.

\section{Analysis}

\subsection{Diagnosis Accuracy and Misclassification Analysis}

\begin{figure}[t]
    \centering
    \includegraphics[width=1\linewidth]{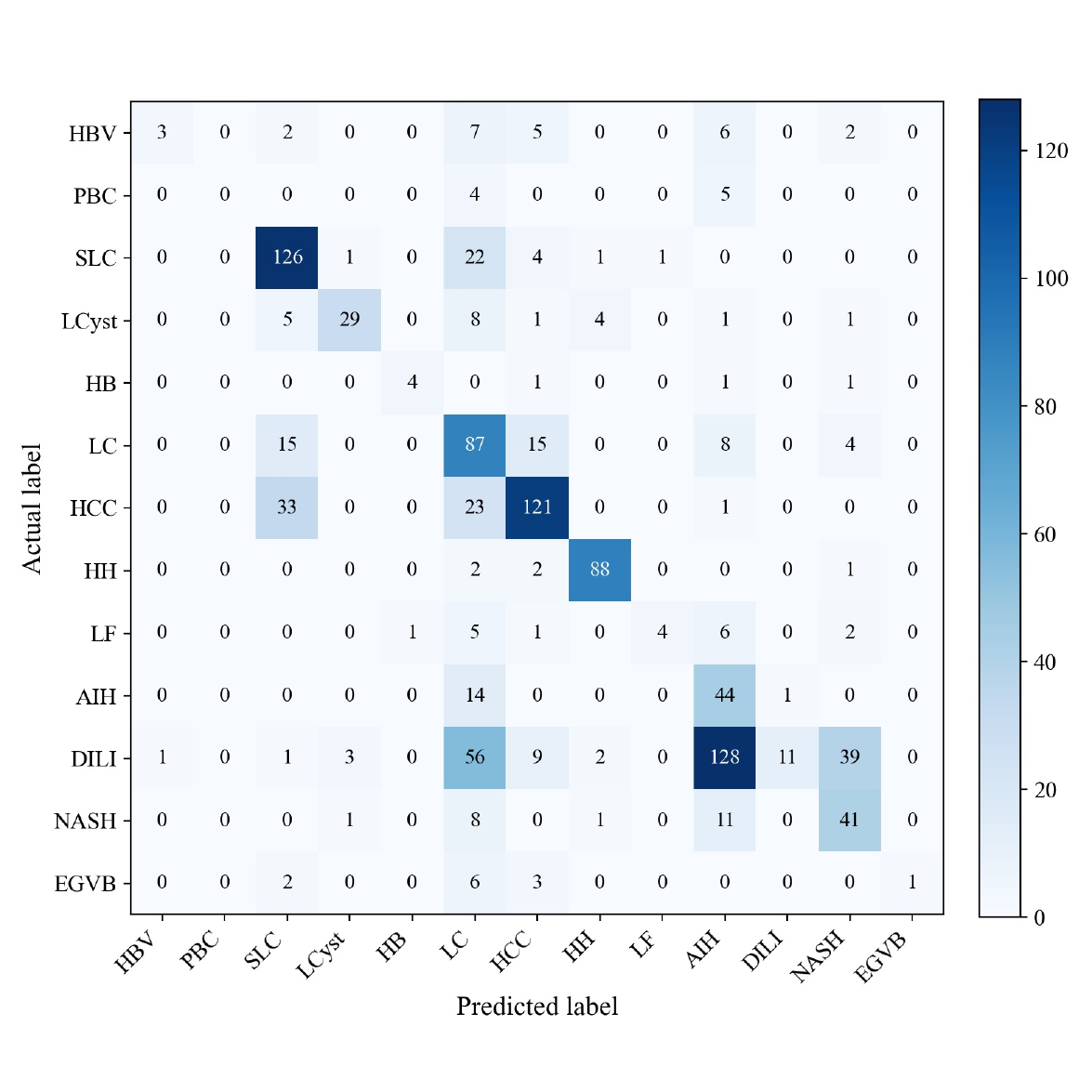}
    \caption{Confusion matrix of \textsc{MedCoRAG} on 13 hepatic disease classes.}
    \label{fig:placeholder}
\end{figure}

Table~\ref{tab:MedCoRAG_disease_metrics_abbr} presents per-disease diagnostic metrics for \textsc{MedCoRAG}, revealing consistently strong performance across a range of hepatic conditions. The model achieves high precision and recall for diseases with distinct clinical or radiological signatures, including hepatic hemangioma, liver cyst, secondary liver cancer, and drug-induced liver injury. The model achieves perfect precision in predicting acute events such as rupture and bleeding of esophagogastric varices, indicating high confidence in critical diagnoses when they are issued. The confusion matrix in Figure~\ref{fig:placeholder} shows that misclassifications primarily occur among clinically related entities—such as cirrhosis-associated complications and cholestatic disorders—reflecting known diagnostic similarities in hepatology. These results demonstrate that \textsc{MedCoRAG} aligns its reasoning with established clinical patterns while maintaining robust accuracy across diverse liver diseases.

\begin{table}[t]
\centering
\caption{Per-disease diagnostic metrics (\%) of \textsc{MedCoRAG}.}
\label{tab:MedCoRAG_disease_metrics_abbr}
\begin{tabular}{@{}l c c c c@{}}
\toprule
\textbf{Disease} & \textbf{Recall} & \textbf{Precision} & \textbf{F1-score} & \textbf{F0.5-score} \\
\midrule
HBV  & 52.00 & 76.47 & 61.90 & 55.56 \\
PBC  & 55.56 & 20.83 & 30.30 & 41.67 \\
SLC  & 85.16 & 92.31 & 88.59 & 86.50 \\
LCyst& 87.76 & 93.48 & 90.53 & 88.84 \\
HB   & 85.71 & 85.71 & 85.71 & 85.71 \\
LC   & 67.44 & 62.59 & 64.93 & 66.41 \\
HCC  & 90.45 & 74.19 & 81.52 & 86.65 \\
HH   & 94.62 & 94.62 & 94.62 & 94.62 \\
LF   & 15.79 & 75.00 & 26.09 & 18.75 \\
AIH  & 69.49 & 54.67 & 61.19 & 65.92 \\
DILI & 82.00 & 88.74 & 85.24 & 83.27 \\
NASH & 69.35 & 87.76 & 77.48 & 72.39 \\
EGVB & 16.67 & 100.00 & 28.57 & 20.00 \\
\bottomrule
\end{tabular}
\end{table}

\subsection{Diagnostic Complexity and Reasoning Patterns}

\begin{figure}[t]
    \centering
    \includegraphics[width=1\linewidth]{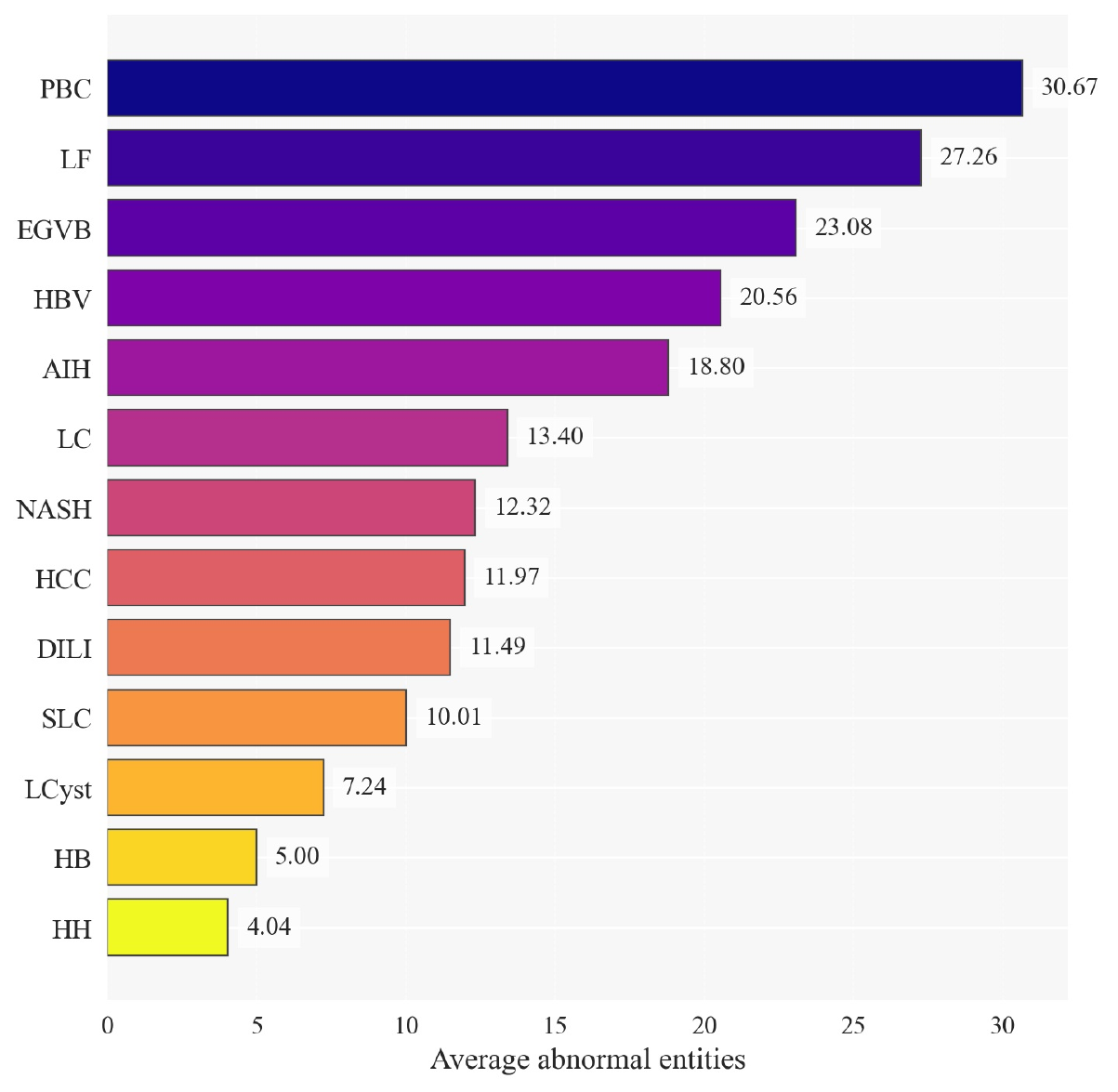}
    \caption{Average number of abnormal entities per case across different hepatic diseases. Higher values indicate more complex clinical presentations.}
    \label{fig:abnormal_entities}
\end{figure}

\begin{figure}[t]
    \centering
    \includegraphics[width=1\linewidth]{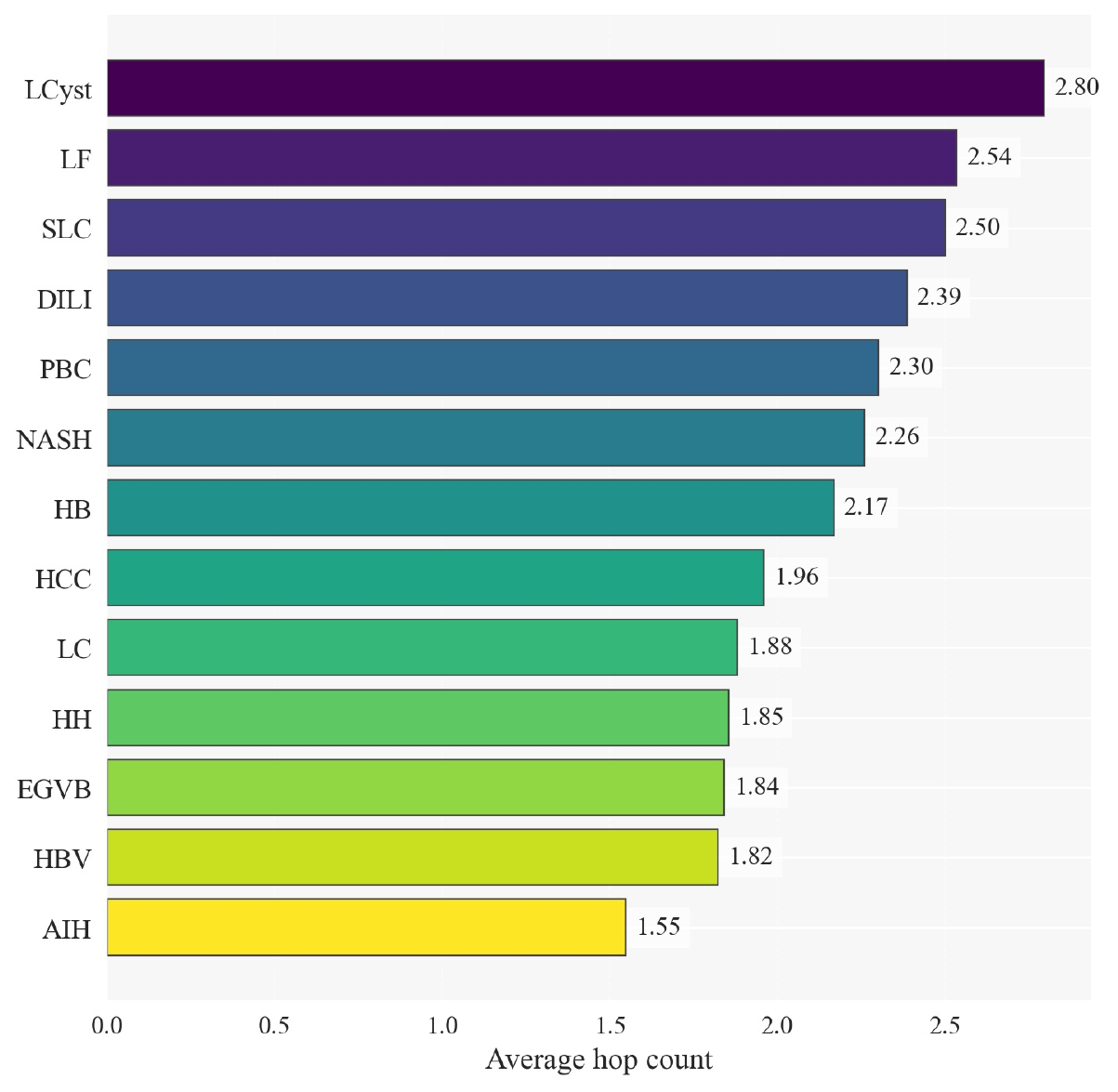}
    \caption{Average number of hops in knowledge graph paths used during diagnosis. Higher values reflect greater reasoning complexity.}
    \label{fig:hop_count}
\end{figure}

We analyze how \textsc{MedCoRAG} adapts to varying diagnostic demands by examining clinical complexity and reasoning depth across diseases. As shown in Figure~\ref{fig:abnormal_entities}, conditions like PBC and LF present with the highest numbers of abnormal entities, reflecting intricate clinical profiles, while HH and HB involve markedly simpler presentations. Correspondingly, Figure~\ref{fig:hop_count} reveals that diagnoses such as LCyst, LF, DILI, and SLC rely on longer knowledge graph reasoning paths, indicating active integration of multi-hop evidence from guidelines and structured medical knowledge. Together, these results illustrate that \textsc{MedCoRAG} tailors its inference process to the inherent complexity of each case, leveraging deeper reasoning where clinical ambiguity is greatest.

\subsection{Ablation Study}
\label{ssub:ablation}
\begin{table}[t]
\small
\centering
\caption{Ablation study results (\%). Best in each column is \textbf{bolded}. 
CG: Clinical Guide integration; 
MA: Multi-Agent deliberation; 
KG: Knowledge Graph grounding; 
TD: Teacher Distillation.}
\label{tab:ablation}
\begin{tabular}{@{}l c c c c@{}}
\toprule
\textbf{Variant} & 
\textbf{Precision} & 
\textbf{Recall} & 
\textbf{F1-score} & 
\textbf{F0.5-score} \\
\midrule
w/o CG & 78.08 & 72.75 & 73.43 & 72.62 \\
w/o MA & 76.30 & 69.72 & 69.70 & 69.14 \\
w/o KG & 77.96 & 73.14 & 73.81 & 73.03 \\
w/o TD & 78.20 & 73.71 & 74.48 & 73.73 \\
w/o KG \& CG & 74.47 & 68.10 & 67.86 & 67.37 \\
w/o TD \& MA & 76.83 & 68.74 & 69.93 & 68.62 \\
w/o KG \& CG \& MA & 69.30 & 57.50 & 55.32 & 55.62 \\
\midrule
\textsc{MedCoRAG} & \textbf{81.32} & \textbf{79.18} & \textbf{79.12} & \textbf{78.99} \\
\bottomrule
\end{tabular}
\end{table}

We assess the contribution of each component in \textsc{MedCoRAG} to diagnostic performance (see Table~\ref{tab:ablation}) by evaluating several ablated variants:  
(1) w/o CG (without Clinical Guide integration), where agents do not access structured clinical guides during reasoning;  
(2) w/o MA (without Multi-Agent deliberation), where diagnosis is generated by a single agent without collaborative discussion;  
(3) w/o KG (without Knowledge Graph grounding), where disease-entity relationships from the medical knowledge graph are excluded;  
(4) w/o TD (without Teacher Distillation), where the student model operates without guidance from the teacher-based reasoning process;  
as well as combined ablations, including w/o KG \& CG, w/o TD \& MA, and w/o KG \& CG \& MA.  

The results show that removing any single component leads to a performance drop across all metrics, with the most substantial decline observed when both knowledge sources (KG and CG) and multi-agent collaboration (MA) are absent (F1 drops to 55.32). Notably, disabling multi-agent deliberation (w/o MA) reduces F1 by 9.42 points compared to the full model, indicating that agent interaction plays a critical role in refining diagnostic hypotheses. Similarly, omitting clinical guides (w/o CG) or the knowledge graph (w/o KG) results in F1 scores below 74, suggesting that structured external evidence is essential for accurate reasoning. The relatively stronger performance of w/o TD (F1: 74.48) implies that while teacher-guided distillation further enhances performance, the core gains stem from the integration of authoritative knowledge and collaborative agent dynamics.

\subsection{Deployment Cost and Efficiency}
\label{ssub:cost}

The one-time cost to construct the teacher-generated training dataset using Qwen3-Max is \$24.53, with no recurring expenses thereafter. At inference time, \textsc{MedCoRAG} exhibits predictable latency: cases requiring multi-agent collaborative reasoning take an average of 33.36 seconds, while simpler cases handled by a single generalist agent complete in just 9.95 seconds.

\subsection{Case Study: How Architecture Enables Expert-Level Diagnosis}
We present a representative case that highlights the advantages of \textsc{MedCoRAG}'s modular design. A 48-year-old female presents with persistent fatigue, pruritus, and jaundice. Lab findings reveal markedly elevated ALP (340 U/L), GGT (280 U/L), and IgM (3.8 g/L), with normal IgG levels. Abdominal ultrasound shows no biliary obstruction but mild hepatomegaly. She denies alcohol use, recent medication changes, or known viral hepatitis exposure.

This presentation is diagnostically challenging due to overlap among PBC, AIH, and DILI. A standard language model might favor AIH or DILI based on fatigue and elevated transaminases, overlooking the cholestatic pattern.

In contrast, \textsc{MedCoRAG} leverages its full architecture for precise reasoning. First, abnormal-entity detection identifies $\mathrm{ALP\_elevated}$, $\mathrm{GGT\_elevated}$, $\mathrm{IgM\_elevated}$, and $\mathrm{pruritus}$, which collectively trigger dynamic routing to the Autoimmune Hepatology Agent while suppressing irrelevant specialists such as Virology or Oncology.

The activated agent then retrieves a clinical guideline excerpt from EASL stating that ``persistent cholestasis with isolated IgM elevation in middle-aged women is highly suggestive of PBC, even in the absence of anti-mitochondrial antibodies.'' Concurrently, the knowledge graph yields a coherent 2-hop path: $\mathrm{pruritus} \rightarrow \mathrm{PBC} \rightarrow \mathrm{IgM\_elevated}$, linking symptoms to serological markers through established disease semantics.

Multi-agent collaboration further refines the diagnosis: the Immunology Agent notes that normal IgG levels argue against typical AIH, while the Hepatology Generalist cross-validates the cholestatic enzyme profile and absence of drug exposure to rule out DILI. The system converges on a high-confidence diagnosis of $\mathrm{PBC}$, accompanied by a traceable rationale and a recommendation for anti-mitochondrial antibody testing.

This case illustrates how \textsc{MedCoRAG}'s components synergize: abnormality-driven routing ensures clinical relevance; hybrid retrieval grounds reasoning in both guidelines and structured knowledge; and multi-agent deliberation enables nuanced differential analysis---all without reliance on invasive findings. This integrated workflow underpins the framework’s ability to resolve ambiguous cases that mimic expert clinical judgment.

\subsection{Limitations and Future Work}
\label{subsec:limitations_future}

\textsc{MedCoRAG} demonstrates strong diagnostic performance and interpretable reasoning on hepatic cases from MIMIC-IV, yet it has several limitations. The current implementation processes only a single clinical snapshot, lacking modeling of longitudinal signals such as lab trends or imaging evolution. It also depends on UMLS-aligned entities and static guidelines, which can be sensitive to ambiguity in real-world clinical notes. Moreover, all evaluations are retrospective and have not been validated in live clinical workflows. To address these issues, future work will incorporate temporal EHR modeling for time-aware diagnosis, extend the agent framework to broader clinical domains beyond hepatology, enhance robustness to unstructured text through improved natural language understanding, develop lightweight mechanisms for efficient deployment, and conduct prospective studies with clinical partners to assess real-world impact on decision support and EHR integration.

\section{Conclusion}
\label{sec:conclusion}
We propose \textsc{MedCoRAG}, a hybrid retrieval-augmented generation and multi-agent framework for interpretable hepatic disease diagnosis. The method constructs a patient-specific evidence package by jointly retrieving and pruning paths from a medical knowledge graph and excerpts from clinical practice guidelines. A router agent dynamically activates relevant specialist agents based on abnormal clinical findings, enabling iterative, role-aware deliberation over the shared evidence packet. Consensus is formed through holistic adjudication by a generalist agent, yielding a single, traceable diagnosis grounded in both structured knowledge and expert guidance. Evaluated on real-world hepatic cases from MIMIC-IV, \textsc{MedCoRAG} outperforms a wide range of baselines, demonstrating superior diagnostic performance and reasoning explainability. This work represents a step toward more transparent, evidence-grounded, and clinically aligned AI for trustworthy medical decision support.

\section*{CRediT authorship contribution statement}
\textbf{Zheng Li}: Conceptualization, Data curation, Formal analysis, Investigation, Methodology, Project administration, Validation, Visualization, Writing – original draft.  
\textbf{Jiayi Xu}: Investigation, Writing – review \& editing. 
\textbf{Zhikai Hu}: Investigation, Writing – review \& editing.
\textbf{Hechang Chen}: Supervision, Validation, Writing – review \& editing.
\textbf{Lele Cong}: Supervision,  Validation, Writing – review \& editing.
\textbf{Yunyun Wang}: Formal analysis, Project administration, Supervision, Validation, Writing – review \& editing.
\textbf{Shuchao Pang}: Formal analysis, Funding acquisition, Project administration, Supervision, Writing – review \& editing.

\section*{Declaration of competing interest}
The authors declare that they have no known competing financial interests or personal relationships that could have appeared to influence the work reported in this paper.

\section*{Data and code availability}
The MIMIC-IV database is publicly available through PhysioNet at: \url{https://physionet.org/content/mimiciv/3.1}.  The source code will be publicly available after acceptance.

\section*{Acknowledgements}
This work is supported by the National Natural Science Foundation of China (Grant No.62206128), National Key Research and Development Program of China (Grant No.2023YFB2703900)

\bibliographystyle{elsarticle-num}
\bibliography{reference}

\end{document}